\documentclass[letterpaper]{article} % DO NOT CHANGE THIS
\usepackage{aaai20}  % DO NOT CHANGE THIS

\usepackage{times}  % DO NOT CHANGE THIS
\usepackage{helvet} % DO NOT CHANGE THIS
\usepackage{courier}  % DO NOT CHANGE THIS
\usepackage[hyphens]{url}  % DO NOT CHANGE THIS
\usepackage{graphicx} % DO NOT CHANGE THIS
\usepackage{graphicx}  % DO NOT CHANGE THIS
\usepackage{caption}
\usepackage{subcaption}
\usepackage{amsmath}
\usepackage{booktabs}
\usepackage{appendix}
\usepackage{multirow}
\usepackage{float}
\usepackage{color}   %% to be removed later
\usepackage{titlesec}
\usepackage{setspace}
\usepackage{url}
\usepackage{amsmath}
\usepackage{xcolor}
\usepackage{breqn}
\usepackage{bm}
\usepackage{amsfonts}
\usepackage{mathtools}
\newcommand{\defeq}{\vcentcolon=}
\newtheorem{problem}{Problem}
\newtheorem{definition}{Definition}
\newtheorem{example}{Example}
\newtheorem{theorem}{Theorem}
\newtheorem{lemma}{Lemma}
\newtheorem{corollary}{Corollary}
\newtheorem{assumption}{Assumption}
\newenvironment{proof}{\paragraph{Proof:}}{\hfill$\square$}

\title{Dis-entangling Mixture of Interventions on a Causal Bayesian Network Using Aggregate Observations}

 \author{
 Gaurav Sinha\textsuperscript{1}, 
 Ayush Chauhan\textsuperscript{2}, 
 Aurghya Maiti\textsuperscript{3}\thanks{The author is at IIT Kharagpur currently. Email: aurghya@iitkgp.ac.in}, 
 Naman Poddar\textsuperscript{4}\thanks{The author is at IIT Guwahati currently. Email: namanpoddar@iitg.ac.in}, 
 Pulkit Goel\textsuperscript{5}\thanks{The author is at IIT Delhi currently. Email: goel.pulkit03@gmail.com}  \\
 \textsuperscript{1}gasinha@adobe.com, 
 \textsuperscript{2}ayuchauh@adobe.com, 
 \textsuperscript{3}aumaiti@adobe.com, 
 \textsuperscript{4}npoddar@adobe.com, 
 \textsuperscript{5}pulgoel@adobe.com   \\
 Adobe Research, Bangalore
 }

\begin{document}

\maketitle

% \titlespacing*{\section}{0pt}{0.5\baselineskip}{0.2\baselineskip}
% \titlespacing*{\subsection}{0pt}{0.1\baselineskip}{0.1\baselineskip}
% \titlespacing*{\paragraph}{0pt}{0.1\baselineskip}{0.1\baselineskip}

% \twocolumn[

% \aistatstitle{Dis-entangling Mixture of Interventions on a Causal Bayesian Network Using Aggregate Observations}

% % \aistatsauthor{Anonymous}
% % \vspace{2em}

% % \aistatsauthor{Gaurav Sinha \And Ayush Chauhan}

% % \aistatsaddress{Adobe Research\\Bangalore\\gasinha@adobe.com \And Adobe Research\\Bangalore\\ayuchauh@adobe.com}

% % \aistatsauthor{Aurghya Maiti\thanks{The author is at IIT Kharagpur currently. Email: aurghya@iitkgp.ac.in} \And Pulkit Goel\thanks{The author is at IIT Delhi currently. Email: goel.pulkit03@gmail.com} \And Naman Poddar\thanks{The author is at IIT Guwahati currently. Email: namanpoddar@iitg.ac.in}}

% % \aistatsaddress{Adobe Research\\Bangalore\\aumaiti@adobe.com \And Adobe Research\\Bangalore\\pulgoel@adobe.com \And Adobe Research\\Bangalore\\npoddar@adobe.com}

% ]

\begin{abstract}
We study the problem of separating a mixture of distributions, all of which come from interventions on a known causal bayesian network. Given oracle access to marginals of all distributions resulting from interventions on the network, and estimates of marginals from the mixture distribution, we want to recover the mixing proportions of different mixture components. 

We show that in the worst case, mixing proportions cannot be identified using marginals only. If exact marginals of the mixture distribution were known, under a simple assumption of excluding a few distributions from the mixture, we show that the mixing proportions become identifiable. Our identifiability proof is constructive and gives an efficient algorithm recovering the mixing proportions exactly. When exact marginals are not available, we design an optimization framework to estimate the mixing proportions.

Our problem is motivated from a real-world scenario of an e-commerce business, where multiple interventions occur at a given time, leading to deviations in expected metrics. 

%The mixing proportions of each intervention distribution can then be interpreted as their contribution towards the deviation. We've restricted access to marginals in our problem setup in order to capture the situation of an analyst who gets to access only aggregate data from analytics reports and dashboards.

We conduct experiments on the well known publicly available ALARM network and on a proprietary dataset from a large e-commerce company validating the performance of our method.
\end{abstract}

\section{Introduction}

\subsection{Motivation}
\label{subsec:motivation}
Causal Bayesian Networks\footnote{called causal networks in short} (Definition \ref{defn:causal-bayesian-network}) are one of the most popular ways of modelling causal relationships in data \cite{Pearl2009}. Broadly speaking, these are directed acyclic graphs with nodes representing random variables such that
\begin{itemize}
    \item the joint probability factors as a product of conditional probabilities of nodes given their parents, and
    \item directed edges indicate direct causal relationships.
\end{itemize}
%Learning such networks from observational data has been pursued extensively in the last few decades and several algorithms for estimating such graphs exist, eg. PC algorithm \cite{Spirtes2000} or Greedy Equivalence Search (GES) algorithm \cite{Meek1997}, \cite{Chickering2003}. However it should be noted that any algorithm using observational data only can identify the network upto an equivalence relation \cite{Chickering2002} output-ing a partially directed graph in many cases. In these cases, the actual direction of the (learned) undirected edges is not identifiable from observational data only.

Learning causal relationships among variables often requires one to ``intervene'' on the network, where in values of some variables are forcibly fixed. If one has the luxury of performing such interventions, identifiability of directions in learning causal bayesian networks is greatly improved \cite{Eberhardt2005}, \cite{Hauser2012}. 

Therefore, interventions are essential for discovering causal relationships, but that is not the only reason why interventions are performed in the real world. As businesses evolve, they perform multiple interventions at the same time (sometimes unknowingly) to their underlying systems for $(a)$ testing new changes, $(b)$ providing better user experience, $(c)$ reducing costs, $(d)$ customer acquisition, $(e)$ operational convenience etc.

Even though this brings value to the business, it becomes a nuisance for an analyst who is trying to diagnose fluctuations in business metrics by looking at aggregate values in reports. In the next paragraph, we discuss a use case which describes such a situation.

\subsubsection{Motivation from e-commerce analytics}
Consider an e-commerce business which records all click-stream data generated by visitors. Any such data instance comprises of multiple features eg: source page of the click, destination page of the click, whether item was purchased etc. Suppose for easier diagnosis of problems in future, the business has learned underlying causal relationships between these features through rigorous data analysis, A/B tests, prior knowledge etc.

We are concerned with a situation when during a certain time range, an analyst discovers significant shift in expected behaviour of some relevant metrics (eg. number of orders etc.) caused by interventions happening in the system. We would like to help the analyst in attributing credit/blame for these shifts, to the individual interventions. But with a severe limitation of having access only to aggregate data for all feature values. The reason for this limitation is that analytics reports and dashboards only reveal aggregate numbers and that is all our analyst can use.

To correctly attribute, we propose modelling distribution of the deviated data as a mixture of distributions resulting from interventions on the system. The mixing proportion of each intervention in the mixture can be seen as the credit/blame attributed to it. 

This motivates us to design an algorithm which uses aggregate data to dis-entangle a mixture of interventions and recover mixing proportions.

\subsubsection{Research Problem} 
%The focus of our paper is not on learning the causal network but instead on dis-entangling data coming from a mixture of interventions on such a network. More precisely, in our setting, every data sample comes from a mixture of several interventional distributions (see definition \ref{defn:intervention}). For simplicity we only consider perfect interventions\footnote{However, the mixture could contain many such perfect interventions and samples could be coming from any of them.} (see discussion following definition \ref{defn:intervention}). 

%Assume we are given:
%\begin{itemize}
%    \item Oracle access to all marginals (of distributions) resulting from perfect interventions on some underlying causal network (see definition \ref{defn:intervention}), and
%    \item Estimates of marginal probabilities of the mixture.
%\end{itemize}
%  We discuss the problem of dis-entangling the mixture i.e. identifying the proportion of each intervention in the mixture. We believe (after extensive literature survey) our work to be the first one to address this problem. In the next paragraph we formally define the problem. Please refer to section \ref{subsection:notation-and-preliminaries} for a more formal definition of the mixture model.
We define our research problem below. Please see section \ref{subsection:notation-and-preliminaries} and appendix \ref{appendix:preliminaries} for any definitions needed.

Let $[n]$ denote the set $\{1,\ldots,n\}$ and $\mathcal{M}$ be a causal bayesian network on ${\bf X} = (X_1,\ldots, X_n)$ such that for every $i\in [n]$, $X_i$ is a discrete random variables taking values in set $C_i$. Let $P$ denote the joint probability distribution of ${\bf X}$ and $P_{i,\alpha}$  denote probability distributions obtained as a result of intervention (Definition \ref{defn:intervention} in Appendix \ref{appendix:preliminaries}) setting $X_i = \alpha$ ($i\in [n], \alpha \in C_i$) in $\mathcal{M}$. Consider a mixture model with distribution $P_{mix}$ such that any value sampled from $P_{mix}$ comes from distribution $P$ with probability $\pi_{\phi}$ and from  $P_{i,\alpha}$ with probability $\pi_{i,\alpha}$. Given access to:
\begin{enumerate}
    \item Directed acyclic graph representing $\mathcal{M}$.
    \item Marginals $P(X_j = \beta)$ and $P_{i,\alpha}(X_j = \beta)$ for all $i,j\in [n]$ and $\alpha \in C_i, \beta\in C_j$.
    \item Marginals (or good estimates of) $P_{mix}(X_j = \beta)$ for all $j\in [n], \beta \in C_j$. 
\end{enumerate}
\noindent
Design an algorithm to recover mixing proportions $\pi_{i,\alpha}$ ($i\in [n], \alpha\in C_i$)\footnote{This also gives us $\pi_\phi = 1-\sum\limits_{i\in[n]}\sum\limits_{\alpha\in C_i}\pi_{i,\alpha}$}.

\subsubsection{Caveat and assumptions} While attempting to solve the above problem, the first difficulty we encounter is whether the problem even has a ``well defined'' solution. In other words whether the mixing proportions $\pi_{i,\alpha}$'s can be uniquely identified using the network $\mathcal{M}$ and marginal probabilities $P(X_j = \beta), P_{i,\alpha}(X_j = \beta)$ and $P_{mix}(X_j = \beta)$ for $i,j\in [n]$ and $\alpha \in C_i, \beta\in C_j$.

Using a simple construction in Example \ref{example:ex1}, we show that in general the $\pi_{i,\alpha}$'s are not identifiable. Thus in the worst case there is no hope of recovering the mixing proportions, even if we know all marginals exactly. 

We then go on to put mild assumptions that will help us get over this hurdle. We establish that if for every node $X_i$ there is some category $\alpha_i$ ($\alpha_i$ not known in advance)
such that the intervention setting $X_i=\alpha_i$ does not appear in the mixture
then the $\pi_{i,\alpha}$'s can be identified by solving a sequence of linear systems \footnote{whose parameters can be computed using oracle queries.}. Note that this still allows as many as $\sum\limits_{i=1}^n (|C_i|-1)+1$ distributions contributing to the mixture.

\subsection{Our contributions}
Here are the main contributions of this work:
\begin{itemize}
    \item We define a mixture model of distributions resulting from interventions on a causal bayesian network motivated by the practical scenario of analyzing shifts in 
    web-metrics. To the best of our knowledge, our paper is the first one to solve this particular problem and apply it to a real world scenario. \item Using a simple construction in Example \ref{example:ex1}, we show that the mixing proportions are not identifiable from data in general and therefore some assumptions are needed for identifiability.
    \item We show that under a mild assumption (Assumption \ref{assumption:identifiability-assumption}), the mixing proportions become identifiable (Theorem \ref{theorem:identifiability}). We also give arguments to support the validity of this assumption in practice. 
    \item If marginals of mixture distribution are known, our proof of Theorem \ref{theorem:identifiability} can be used as an efficient algorithm to recover mixing proportions exactly.
    \item Since in our problem setup only estimates of marginals for the mixture distribution are available, the above algorithm does not work. To deal with this, we design an optimization approach (called Deviations in Marginals Method or DIMM) rooted in proof of Theorem \ref{theorem:identifiability} and solve it using sequential quadratic programming \cite{Nocedal2006} implementation provided in the scipy \cite{scipy} library.
    \item We validate DIMM on two datasets. Our first dataset comes from the well known
    ALARM network \cite{beinlich1989alarm} and the second dataset is from a large e-commerce business. 
\end{itemize}
\subsection{Relevant prior work}
\cite{Thiesson1998} is perhaps the most relevant work for this paper. In their case the observed data comes from a mixture of (possibly distinct) DAG (directed acyclic graph) models unlike ours where the mixture components are interventions on the same ground network. Secondly they try to learn structures and mixing proportions at the same time which makes the problem quite hard. They design a heuristic method with no guarantees on correct recovery of mixture components and proportions. On the other hand, we give a proof for identifiability of mixing proportions under a mild assumption (Assumption \ref{assumption:identifiability-assumption}).

Mixtures of interventions have also appeared in other studies. In section $6$ of \cite{Korb2004}, ``partially effective'' interventions were defined as mixtures of the traditional $do()$ interventions and the base network. These were also called ``unreliable interventions'' in \cite{Eaton2007}.
In this model, there is an effectiveness probability with which an intervention succeeds thereby leading to a mixture of two distributions. Our paper generalizes such mixtures into a mixture of many distributions resulting from interventions.

%Causal discovery from interventions has been a very active area of research recently. \cite{Eberhardt2005} gave tight bounds on the number of interventions necessary and sufficient to discover all causal relations among $N$ random variables. \cite{Eaton2007} applied dynamic programming structure learning algorithm of \cite{Koivisto2004} to the case when data is obtained from experiments. They consider different intervention models including the $do()$ interventions and generalizations such as ``uncertain interventions''.

Our problem setup includes situation where a very few number of interventions could have participated in the mixture. Our algorithm can be used to find which ones were present by just considering non-zero mixing proportions. So it also provides a method to identify which interventions are part of the mixture, an information that is a priori not known.
There are works which partly deal with this under similar/more general definitions of interventions. For example, \cite{Tian2001} proposed a model of interventions which they call ``mechanism change'', wherein the intervention does not set the value of a node, but changes the conditional distribution of the node given its parents. In section $4.4$, they discuss the case where ``focal variables'' (i.e. the ones directly affected by mechanism change) are not known in advance and describe conditions under which they can be identified from data (to some extent).
In a related line of work \cite{Eaton2007} designed causal discovery algorithms for ``uncertain interventions'' in which multiple interventions occur on the network with each intervention influencing multiple nodes, therefore the exact targets of interventions are not known. In their setup, it was assumed that for each data sample, some of the interventions were active, indicating what mechanism change occurred (at all affected nodes). We encourage the interested reader to find more details in their paper. 

\subsection{Preliminaries}
\label{subsection:notation-and-preliminaries}
Throughout this paper for every $n\in \mathbb{N}$ the set $\{1,\ldots,n\}$ is denoted by $[n]$. We also use definitions of Bayesian Networks and Interventions which are common knowledge in the causality literature and can be found in Appendix \ref{appendix:preliminaries} or in the cited literature. Below we define Causal Bayesian Networks and a mixture model of distributions resulting from interventions. 

Loosely speaking, a causal bayesian network is a bayesian network with edges indicating direct causal relationships, enabling an elegant process of intervention on the nodes. However, it can be formally defined with respect to the set of all interventional distributions as follows.

\begin{definition}[Causal Bayesian Network, $1.3.1$, \cite{Pearl2009}]
\label{defn:causal-bayesian-network}
Let $P$ be a probability distribution for random variable ${\bf X} = (X_1,\ldots,X_n)$. Let $C_i$ be the set of values taken by $X_i$. For any subset $S\subset [n]$ consider the interventional distribution 

\[
P_{x_S} \defeq P({\bf X} | do({\bf X_S} = x_S))
\]
where $x_S = (x_s)_{s\in S} \in \prod\limits_{s\in S}C_s$. Define ${\bf P_{\star}}$ to be the collection 
\[
{\bf P_{\star}} = \{P_{x_S} : S\subset [n], x_S\in \prod\limits_{s\in S}C_s\}
\]
A bayesian network $\mathcal{M} = (G,\mu)$ (on ${\bf X}$)  is called a causal bayesian network compatible with ${\bf P_{\star}}$ if and only if the following three conditions hold for every $P_{x_S}\in {\bf P_{\star}}$.
\begin{enumerate}
    \item $P_{x_S}$ is Markov relative to $G$;
    \item For all $i\in S$ and $y_i$ consistent with ${\bf X_S} = x_S$, 
    \[
    P_{x_S}(X_i = y_i)=1;
    \]
    \item For all $i\notin S$ and $pa_i$ consistent with $X_S = x_S$,
    \begin{equation*}
        \begin{split}
            P_{x_S}(X_i = y_i | Pa(X_i) = pa_i) = \\ P(X_i = y_i | Pa(X_i) = pa_i)
        \end{split}
    \end{equation*}
\end{enumerate}
\end{definition}

The above definition constraints the following truncated factorization for every $P_{x_S}\in {\bf P_{\star}}$,
\[
P_{x_S}(y) = \prod\limits_{i\notin S}P(y_i | pa_i) 
\]
for all $y = (y_1,\ldots,y_n)$ consistent with ${\bf X_S}=x_S$. 

Given this definition, we are now ready to define a mixture model of perfect interventions (see Definition \ref{defn:intervention} in Appendix \ref{appendix:preliminaries}) which is the central object we analyze in this paper.

\begin{definition}[Mixture of perfect interventions]
\label{defn:mixture-model}
Let $\mathcal{M} = (G,\mu)$ be a causal bayesian network compatiable with ${\bf P_{\star}}$ as given in Definition \ref{defn:causal-bayesian-network} in Appendix \ref{appendix:preliminaries}.
In this work, we deal with a mixture of perfect interventions. Thus, distributions corresponding to these interventions are contained in the set 
\[
{\bf Q} = \{P_{x_S}\in {\bf P_{\star}} : |S|\leq 1, x_S\in \prod\limits_{i\in S}C_i\}
\] 

For the rest of this paper, we only focus on categorical random variables $X_i, i\in [n]$ taking values in finite sets $C_i$ (respectively). For better exposition, we simplify the notation for categorical variables. When no intervention occurs i.e. $S=\phi$, we call the distribution $P_{\phi}$\footnote{this is same as the ground distribution.}, when intervention occurs on exactly one node, say $S = \{i\}$, setting $X_i$ to some $\alpha\in C_i$, we call the distribution $P_{i,\alpha}$. Therefore,
\[
{\bf Q} = \{P_{i,\alpha} : i\in [n], \alpha\in C_i\}\cup \{P_\phi\}
\]
is a finite set of size $\sum\limits_{i=1}^n |C_i| + 1$. We define a mixture of distributions from ${\bf Q}$ using a latent variable $Z$ which can take values in the set $\{(i,\alpha) : i\in [n], \alpha \in C_i\}\cup \{\phi\}$. Using this we can define the joint distribution $P_{mix}$ of the mixture as:
\[
P_{mix}({\bf X}) = \sum\limits_{P_s\in {\bf Q}}P_s({\bf X})Pr[Z=s]
\]
\end{definition}

\subsection{Organization of the paper}
In section \ref{section:problem-statement} we define the problem formally, show that identifiability does not hold in general, clearly define assumptions needed and give an identifiability proof under these assumptions. Using the proof idea from section \ref{section:problem-statement}, in section \ref{section:optimization-formulation-and-main-algorithm} we setup an optimization problem whose solutions will give us estimates of the desired mixing proportions. In section \ref{section:data-and-preprocess} we provide more details on the data used and preparation of the oracle $\mathcal{O}$. In section \ref{section:Expts-and-results} we describe the experimental setup and show results for the experiments mentioned. Finally in section \ref{section:conclusions}, we conclude the paper with directions for future work.
\section{Problem statement, assumptions and identifiability algorithm}
\label{section:problem-statement}

Let $\mathcal{M} = (G,\mu)$ be a causal bayesian network on variables ${\bf X} = (X_1,\ldots,X_n)$ (with distribution $P$) compatible with the set of distributions ${\bf P_{\star}}$ (as in Definition \ref{defn:causal-bayesian-network}). Further assume there is a mixture model of perfect interventions as defined in section \ref{subsection:notation-and-preliminaries} with distribution $P_{mix}$. Therefore,
\[
P_{mix}({\bf X}) = \sum\limits_{P_s\in {\bf Q}}P_s({\bf X})Pr[Z=s]
\]
where $ {\bf Q} = \{P_{i,\alpha} : i\in [n], \alpha\in C_i\}\cup \{P_\phi\}$, is the set of distributions resulting from perfect interventions. Recall from section \ref{subsection:notation-and-preliminaries} that $P_{i,\alpha}$ is the distribution in ${\bf P_{\star}}$ resulting from the intervention setting $X_i$ to category $\alpha\in C_i$ and $P_{\phi}$ is the distribution when no intervention has occurred (i.e. same as the ground distribution $P$ of ${\bf X}$).

Define $\pi_{i,\alpha} = Pr[Z=(i,\alpha)]$ and $\pi_\phi = Pr[Z=\phi]$. Using this notation, the mixture distribution becomes,
\begin{equation}
\label{equation:perfect-intervention-mixture}
P_{mix}({\bf X}) = \sum\limits_{i\in [n]}\sum\limits_{\alpha\in C_i}P({\bf X} | do(X_i = \alpha))\pi_{i,\alpha} + P({\bf X})\pi_\phi
\end{equation}

We recall our problem statement from section \ref{subsec:motivation}.
\begin{problem}
\label{problem:problem-statement}
Given access to:
\begin{enumerate}
    \item Directed acyclic graph representing $\mathcal{M}$.
    \item Marginals $P(X_j = \beta)$ and $P_{i,\alpha}(X_j = \beta)$ for all $i,j\in [n]$ and $\alpha \in C_i, \beta\in C_j$.
    \item Marginals (or good estimates of) $P_{mix}(X_j = \beta)$ for all $j\in [n], \beta \in C_j$. 
\end{enumerate}
\noindent
Design an algorithm to recover mixing proportions $\pi_{i,\alpha}$ ($i\in [n], \alpha\in C_i$)\footnote{This also gives us $\pi_\phi = 1-\sum\limits_{i\in[n]}\sum\limits_{\alpha\in C_i}\pi_{i,\alpha}$}.
\end{problem}

The first question to address here is if the mixing proportions are even identifiable using marginals, because if they are not it does not make much sense to design any algorithm for the task. In the following subsection we give an example to show that in general they are not identifiable and that will lead us to making reasonable assumptions for identifiability.

\subsection{Identifiability issues}
We illustrate this issue with a uni-variate example for simplicity. This can be generalized to the straight forward multi-variate example where all variables are independent i.e. the network is fully disconnected.
\begin{example}
\label{example:ex1}
Consider a random variable ${\bf X} = (X_1)$ taking values in set $C_1 = \{H,T\}$. Using the notation above, there are three perfect interventions possible on this variable, $P_{1,H}, P_{1,T}$ and $P_\phi$, and the corresponding mixing proportions are $\pi_{1,H}, \pi_{1,T}$ and $\pi_\phi$. Marginals in the mixture can therefore be written as,
\begin{equation*}
    \begin{split}
        P_{mix}(X_1 = x_1) =& P(X_1 = x_1)\pi_\phi +\\& P(X_1 = x_1 | do(X_1 = H))\pi_{1,H}  +\\&  P(X_1 = x_1 | do(X_1 = T))\pi_{1,T} 
    \end{split}
\end{equation*}
for $x_1\in \{H,T\}$. The two equations obtained using $x_1 = H$ and $x_1 = T$ are the only information derivable from the marginals of the mixture distribution. Substituting $P_{mix}(X_1=H)=q$ ($\Rightarrow P_{mix}(X_1=T) = 1-q$) and $P(X_1 = H)=p$ ($\Rightarrow P(X_1=T) = 1-p$) along with $\pi_\phi = 1-\pi_{1,H} - \pi_{1,T}$ in the above equations we get,
\[
q = (1-\pi_{1,H}-\pi_{1,T})\times p + \pi_{1,H},
\]
\[
1-q = (1-\pi_{1,H}-\pi_{1,T})\times (1-p) + \pi_{1,T} 
\]

It's easy to check that both equations are linearly dependent and thus we are left with just one equation $q = (1-\pi_{1,H}-\pi_{1,T})\times p + \pi_{1,H}$ which cannot be solved uniquely for $\pi_{1,H}, \pi_{1,T}$.
\end{example}

Therefore the general problem as described in Problem \ref{problem:problem-statement} is not solvable uniquely for the mixing proportions. The above example generalizes to the fully disconnected networks in a straight forward way implying that there exist multi-variate cases where mixture proportions cannot be identified. To deal with this we make a mild assumption and then show identifiability.

\subsection{Assumptions to ensure identifiability}
In order to show identifiability, we need to show that mixing proportions can be uniquely recovered given the network $\mathcal{M}$ and marginals for $P_{mix}, P$ (which is of bayesian network $\mathcal{M}$) and $P_{i,\alpha}$, $i\in [n]$, $\alpha\in C_i$. Taking marginals with respect to $X_j, j\in [n]$ in equation \ref{equation:perfect-intervention-mixture}, for every $\beta\in C_j$ we get the following equation.

\begin{equation}
\label{equation:perfect-intervention-mixture-marginal}
\begin{split}
  P_{mix}(X_j = \beta) =  &\sum\limits_{i\in [n]}\sum\limits_{\alpha\in C_i} P(X_j=\beta | do(X_i = \alpha)) \pi_{i,\alpha} + \\& P(X_j=\beta)\pi_{\phi} 
\end{split}
\end{equation}

By varying $j\in [n]$ and $\beta\in C_j$, we get a system of linear equations. To identify the mixing proportions, we can try to solve this system, but this won't work directly as illustrated by Example \ref{example:ex1} and thus we make some assumptions.

\begin{assumption}[Identifiability assumptions]
\label{assumption:identifiability-assumption}
For every $i\in [n]$, there exists some $\alpha_i\in C_i$ such that 
\[
\pi_{i,\alpha_i} = Pr[Z=(i,\alpha_i)] = 0. 
\]
\end{assumption}

These assumptions are reasonable from a practical perspective, since it's highly unlikely that any real system will be undergoing all possible perfect interventions for a node at the same time. Also, we want to highlight that we do not fix the $\alpha_i$'s in advance. Just the existence of such $\alpha_i$'s is good enough for proving identifiability which we do next. From here onwards, without loss of generality we assume that $X_1,\ldots,X_n$ are in a topologically sorted order.

\begin{theorem}[Identifiability of mixing proportions]
\label{theorem:identifiability}
Given network $\mathcal{M}$ and marginal probabilities
\[ 
P_{mix}(X_j = \beta), P(X_j = \beta), P(X_j = \beta | do(X_i=\alpha))
\] 
for all $i,j\in [n], \alpha\in C_i, \beta\in C_j$ and that Assumption \ref{assumption:identifiability-assumption} holds. There exists a unique solution to the linear system obtained by varying $j\in [n]$ and $\beta\in C_j$ in equation \ref{equation:perfect-intervention-mixture-marginal}.
\end{theorem}

\begin{proof}
For cleaner exposition we use notation,
\begin{equation*}
    \begin{split}
        & p_{mix}(j,\beta) = P_{mix}(X_j = \beta), \\& p(j,\beta) = P(X_j = \beta), \\& p_{i,\alpha}(j,\beta) = P(X_j = \beta | do(X_i=\alpha)) 
    \end{split}
\end{equation*}
Under this new notation for $j\in [n], \beta\in C_j$, equation \ref{equation:perfect-intervention-mixture-marginal} can be re-written as,

\begin{equation}
\label{equation:perfect-intervention-mixture-marginal-notation}
\begin{split}
  p_{mix}(j, \beta) = \sum\limits_{i\in [n]}\sum\limits_{\alpha\in C_i} p_{i,\alpha}(j, \beta)  \pi_{i,\alpha} + p(j, \beta)\pi_\phi
\end{split}
\end{equation}

The equations can be further simplified using the following two straight-forward consequences from definition of causal bayesian networks (see Definition \ref{defn:causal-bayesian-network}). Let $i,j\in [n]$ with $j\leq i$ and $\alpha\in C_i$, $\beta\in C_j$, then
\[
p_{i,\alpha}(j,\beta) = \begin{cases}
    1 & j=i, \alpha=\beta \\
    0 & j=i, \alpha\neq \beta\\
    0 & j < i
\end{cases}
\]
On applying the above along with substituting $\pi_\phi = 1-\sum\limits_{i\in [n]}\sum\limits_{\alpha\in C_i}\pi_{i,\alpha}$ in equation \ref{equation:perfect-intervention-mixture-marginal-notation} and re-arranging we get,
\begin{equation}
\label{equation:node-equation}
\begin{split}
   & \pi_{j,\beta}(1-p(j,\beta)) - p(j,\beta)\sum\limits_{\substack{\alpha \in C_j \\ \alpha\neq \beta}}\pi_{j,\alpha} =\\ & p_{mix}(j, \beta) - p(j, \beta) - \sum\limits_{i < j}\sum\limits_{\alpha\in C_i} (p_{i,\alpha}(j, \beta) - p(j, \beta) ) \pi_{i,\alpha}  
\end{split}
\end{equation}
Fix some $j\in [n]$ and let $C_j = \{\beta_1, \ldots, \beta_K\}$. By varying $\beta\in C_j$ in equation \ref{equation:node-equation} we know that the vector $(\pi_{j,\beta_1}, \ldots,\pi_{j,\beta_K})$ is a solution to the following linear system in variables $\{x_1,\ldots,x_K\}$.
\[
  \begin{bmatrix}
    1-a_1 & -a_1 & . & . & -a_1 \\
    -a_2 & 1-a_2 & . & . & -a_2\\
    . & . & . & . & .\\
    . & . & . & . & .\\
    -a_K & -a_K & . & . & 1-a_K
    
  \end{bmatrix}
  \begin{bmatrix}
  x_1\\
  .\\
  .\\
  .\\
  x_K
  \end{bmatrix}
   = 
   \begin{bmatrix}
  b_1\\
  .\\
  .\\
  .\\
  b_K
  \end{bmatrix}
\]
where for every $k\in [K]$, we used the substitution,
\begin{equation*}
    \begin{split}
        a_k = \hspace{0.5em} & p(j, \beta_k), \text{ and } \\
        b_k = \hspace{0.5em} & p_{mix}(j, \beta_k) - p(j, \beta_k) - \\ 
        & \sum\limits_{\substack{i < j \\ \alpha\in C_i}} (p_{i,\alpha}(j, \beta_k) - p(j, \beta_k) ) \pi_{i,\alpha}
    \end{split}
\end{equation*}

Denoting the above matrix as ${\bf A}$, vector $(b_1,\ldots,b_K)^T$ as ${\bf b}$ and vector $(x_1,\ldots,x_K)^T$ of variables as ${\bf x}$, our system of equations has the following matrix form:
\begin{equation}
\label{equation:matrix-form}
    {\bf Ax = b}
\end{equation}

Note that ${\bf b}$ will contain terms involving $\pi_{i,\alpha}$ with  $i < j$ and $\alpha\in C_i$. We are going to be building an inductive proof so for the sake of rest of the proof let's assume these have been identified and therefore ${\bf b}$ is a vector of scalars.

We observe that the sum of all rows of ${\bf A}$ is the zero vector and therefore the rows form a dependent set implying that $rank({\bf A}) < K$.

Since $X_j$ is a discrete random variable with range $C_j = \{\beta_1,\ldots, \beta_K\}$, without loss of generality we can assume that for all $k\in [K]$, $a_k = p(j, \beta_k)\neq 0$ \footnote{If there is a category with zero probability, it can be ignored throughout the problem.}. Now we try to find general solutions to the linear system in equation \ref{equation:matrix-form}. For $k\in [K-1]$ we can do row operations on the above system as:
\[
R_k\mapsto R_k - \frac{a_k}{a_K}R_k,
\]
where $R_k, k\in [K]$ is the $k^{th}$ row of the linear system in equation \ref{equation:matrix-form}. After these transformations the system of equations becomes,
\[
  \begin{bmatrix}
    1 & 0 & 0 & . & 0 & \frac{-a_1}{a_K} \\
    0 & 1 & 0 & . & 0 & \frac{-a_2}{a_K}\\
    . & . & . & . & .& .\\
    . & . & . & . & .& .\\
    -a_K & -a_K & . & . & . & 1-a_K
    
  \end{bmatrix}
  \begin{bmatrix}
  x_1\\
  .\\
  .\\
  .\\
  x_K
  \end{bmatrix}
   = 
   \begin{bmatrix}
  b_1^\prime\\
  .\\
  .\\
  .\\
  b_K^\prime
  \end{bmatrix}
\]
where $b_k^\prime = b_k - \frac{a_k}{a_K}b_K$ for $k\in [K-1]$ and $b_K^\prime = b_K$. 

It's easy to see that the first $K-1$ rows of the matrix above are independent and therefore this matrix has rank $\geq K-1$. It's obtained from ${\bf A}$ by row operations and thus $K-1\leq rank({\bf A}) < K \Rightarrow rank(A) = K-1$.

The first $K-1$ equations in the above system give us the set of solutions.
\begin{equation}
\mathcal{L} = 
\label{equation:solution-line}
\{(x_1,\ldots,x_K) : x_k = \frac{a_k}{a_K}x_K + b_k^\prime\}
\end{equation}
Equation \ref{equation:solution-line} describes a line $\mathcal{L} \subset \mathbb{R}^K$ characterizing all solutions of equation \ref{equation:matrix-form} (which are infinitely many at the moment). From here onwards we will use Assumption \ref{assumption:identifiability-assumption} which requires that $\mathcal{L}$ intersects $\mathcal{H}_1\cup\ldots\cup\mathcal{H}_K$, where
\[
\mathcal{H}_k \defeq \{x_k = 0\}
\]
is defined as the hyperplane orthogonal to the co-ordinate axis $x_k$ for each $k\in [K]$.

We make the following claims which help us move towards the final step of our proof. For better exposition the proofs of the following have been moved to Appendix \ref{appendix:proofs}.
\begin{lemma}
\label{lemma:intersection}
The following are true:
\begin{enumerate}
    \item For each $k\in [K]$, $\mathcal{L}\cap\mathcal{H}_k$ is a unique vector (say) $p_k \in \mathbb{R}^K$. Given equations for $\mathcal{L}$, vector $p_k$ can be constructed.
    \item The vector $(\pi_{j,\beta_1}, \ldots,\pi_{j,\beta_K})^T \in \mathcal{P} =  \{p_1,\ldots,p_K\}$.
\end{enumerate}
\end{lemma}

\begin{lemma}
\label{lemma:uniqueness}
There exists at-most one $p \in \mathcal{P} = \{p_1,\ldots,p_k\}$, such that $p$ belongs to $\mathbb{R}_{\geq 0}^K$.
\end{lemma}

Since $(\pi_{j,\beta_1}, \ldots,\pi_{j,\beta_K})^T$ is a vector of probabilities, it belongs to $\mathbb{R}_{\geq 0}^K$. It also belongs to $\mathcal{P}$ by Lemma \ref{lemma:intersection} above. Therefore, by Lemma \ref{lemma:uniqueness} it is the only such vector in $\mathcal{P}$ and can easily be found by scanning through $\mathcal{P}$.

Completing the inductive proof from here is straight-forward. Recall that $X_1,\ldots,X_n$ are assumed to be in a topologically sorted order.
\begin{enumerate}
\item {\bf Base case $j=1$ :} Using marginals $p_{mix}(1,\beta)$ and $p(1,\beta)$\footnote{Since $j=1$, no marginal of $P_{i,\alpha}$ is needed as per equation \ref{equation:node-equation}.} for $\beta\in C_1$, we can write the equation of line $\mathcal{L}$ (see equation \ref{equation:solution-line}). Then using construction in Lemma \ref{lemma:intersection}, we can construct the set $\mathcal{P}$. Finally, we can iterate through $\mathcal{P}$ to find the unique point in $\mathbb{R}_{\geq 0}^{|C_1|}$ and identify $(\pi_{j,\beta_1}, \ldots, \pi_{j,\beta_{|C_1|}})$ where $C_1 = \{\beta_1,\ldots, \beta_{|C_1|}\}$.

\item {\bf Induction step $j>1$ :} Using marginals $p_{mix}(j,\beta)$ and $p(j,\beta)$ for $\beta\in C_j$ along with marginals $p_{i,\alpha}(j,\beta)$ and mixing proportions $\pi_{i,\alpha}$\footnote{already computed by induction since $i<j$.} for $i<j$ and $\alpha\in C_i$, we can write the equation of line $\mathcal{L}$ (see equation \ref{equation:solution-line}). Then using construction in Lemma \ref{lemma:intersection}, we can construct the set $\mathcal{P}$. Finally, we can iterate through $\mathcal{P}$ to find the unique point in $\mathbb{R}_{\geq 0}^{|C_j|}$ and identify $(\pi_{j,\beta_1}, \ldots, \pi_{j,\beta_{|C_j|}})$ where $C_j = \{\beta_1,\ldots, \beta_{|C_j|}\}$.
\end{enumerate}
\end{proof}

This proof is clearly constructive and can be easily converted into an algorithm if access to exact marginals of the mixture distribution is given. Since this is pretty straight-forward, we leave details for the reader and summarize the statement in the corollary below.

\begin{corollary}
\label{corollary:exact-proof}
Given access to directed acyclic graph representing causal network $\mathcal{M}$ and access to marginals $P(X_j=\beta), P_{mix}(X_j=\beta)$ and $P_{i,\alpha}(X_j=\beta)$ for all $i,j\in [n]$ and $\alpha\in C_i, \beta\in C_j$, under Assumption \ref{assumption:identifiability-assumption} there exists a deterministic algorithm that runs in time polynomial in the total number of categories $\sum\limits_{j\in [n]}|C_j|$ and outputs the exact values of mixing proportions $\pi_{j,\beta}$ for $j\in [n]$, $\beta\in C_j$. 
\end{corollary}

\section{Optimization formulation and algorithm}
\label{section:optimization-formulation-and-main-algorithm}

Corollary \ref{corollary:exact-proof} needs exact values of marginals for the mixture distribution. As stated in Problem \ref{problem:problem-statement}, we might only get access to good estimates of these marginals. Algorithm given in proof of Theorem \ref{theorem:identifiability}, heavily depends on knowing the exact marginals and therefore will fail in this more general case. To tackle this problem in Appendix \ref{subsection:optimization-setup} we define an optimization problem aimed at finding good estimates to the exact solution. In Appendix \ref{subsection:description-slsqp} we describe an iterative solver for this problem using a sequential quadratic programming approach. Since our optimization problem is inspired from proof of Theorem \ref{theorem:identifiability} which relies on comparing marginals, we call our optimization approach as the Deviations in Marginals Method or DIMM for short.

\section{Data}
\label{section:data-and-preprocess}
The proposed approach is tested by running multiple sets of experiments on two datasets -- 1) a publicly available Bayesian network dataset and 2) a real world proprietary dataset from a large e-commerce company 
\subsection{ALARM data}
\label{subsec:ALARM_data}
ALARM \cite{beinlich1989alarm} is a publicly available causal bayesian network on categorical variables. It is a fully-connected medium-sized network with 37 nodes connected by 46 edges with a maximum in-degree of 4. The total number of categories in the data are 105 with each node taking 2 to 4 categories.

\subsection{E-commerce data}
\label{subsec:e_commerce_data}
The real world dataset used in our experiments is the online click-stream data generated by user interactions on the website of a large e-commerce company. It consists of a total of 32 features varying from user-specific attributes 
%such as city, country, zip, etc. 
to hit-level attributes (eg. choice of browser, hit source, pagename, referrer type, etc). 
%and other visit- related features like browser, os, mobile id, etc. 
However, none of these features can be used to uniquely identify any visitor. Each row of the click-stream records a unique hit on the website generated by user activities such as purchase, product view, cart addition, etc. 

Using click stream data from a time period where metrics were stable, we learn the underlying causal bayesian network $\mathcal{M}$ between the different data features. This is achieved using open-source causal discovery software Tetrad \cite{tetrad} along with prior knowledge about the data. Since our problem statement assumes access to aggregate data only from the mixture (in order to model limitations of an analyst) using an oracle, we construct that below.

\subsection{Preparing access to marginals}
\label{subsec:prepare-oracle}
We compute marginals (as needed in Problem \ref{problem:problem-statement}) from this network in the following way. 
\begin{itemize}
    \item Marginal probabilities of the underlying distribution is calculated one time using standard bayesian network inference algorithms.
    %variable elimination based inference method.
    \item For each perfect intervention, a network is created by breaking edges from parents to the intervened node and fixing value of intervened node (using ``surgery'' as described in \cite{Pearl1998}). Marginal probabilities of distribution resulting from these networks, is calculated one time by using standard bayesian network inference algorithms
    %variable elimination based inference method.
    \item In case of ALARM data, we consider some fixed values of mixing proportions $\pi_{i,\alpha}$'s for $i\in [n], \alpha\in C_i$ (chosen in a generic way). For $m\in \mathbb{N}$ (for multiple values of $m$), we sample $m$ independent instances from the mixture described by the mixing proportions and component distributions (resulting from interventions). The component distributions can easily be obtained by the ``surgery'' method described above. Finally, we return an estimate for marginals by aggregating over the $m$ samples. 
    \item In case of e-commerce data, for $m\in \mathbb{N}$ (for multiple values of $m$), we sample $m$ independent instances from the time period where metrics are unstable and show significant deviation from expected behavior. We return an estimate for marginals by aggregating over the $m$ samples. 
\end{itemize}
\section{Experiments and Results}
\label{section:Expts-and-results}

\subsection{ALARM data experiments}
\label{subsec:ALARM-data-expts}
We create a number of problem instances for experimentation.
\begin{itemize}
    \item Each problem instance corresponds to a choice of $N_{ivn}$ i.e. the number of interventional distributions corresponding to the mixture. For our experiments $N_{ivn}$ is set to values $\{0, 5, 9, 17, 34, 68 \}$.
    \item When $N_{ivn} = 0$, we set $\pi_{\phi}=1$ and $\pi_{i,\alpha} = 0$, for all $i\in [n], \alpha\in C_i$.
    \item When $N_{ivn} \neq 0$, we first randomly choose $\alpha_i\in C_i$ for $i\in [n]$ and set $\pi_{i,\alpha_i} = 0$ to satisfy Assumption \ref{assumption:identifiability-assumption}. Then we set $\pi_{\phi} = 0.2$ and randomly choose positive values for the remaining $\pi_{i,\alpha}$  $\alpha (\neq \alpha_i) \in C_i$ such that
    \[
    \sum\limits_{i=1}^n\sum\limits_{\alpha(\neq \alpha_i)\in C_i} \pi_{i,\alpha} = 0.8,
    \]
\end{itemize}
Each such selection gives us a mixture distribution to work with and is then used by the oracle as defined above.

Each of the problem instances defined above are then solved iteratively using SLSQP (as described in Appendix \ref{subsection:description-slsqp}). For all of these experiments we chose $\lambda = 0.1$ and $\epsilon = 1e-5$ after manual tuning. The metrics used for evaluation are:
\begin{equation*}
    \begin{split}
        & MSE = \frac{1}{N^c}\sum_{i=1}^{n}\sum_{\alpha \in C_i} (\pi_{i,\alpha} - \hat{\pi}_{i,\alpha})^2 \\
        & MAE = \frac{1}{N^c}\sum_{i=1}^{n}\sum_{\alpha \in C_i} \left |\pi_{i,\alpha} - \hat{\pi}_{i,\alpha}\right | \\
        & MABRE = \max_{i \in [n], \alpha \in C_i} \left |\pi_{i,\alpha} - \hat{\pi}_{i,\alpha}\right | \\
        & \Delta(\bm{\Pi}, \bm{\hat{\Pi}}) = \left | Obj(\bm{\Pi}) - Obj(\bm{\hat{\Pi}})\right |
    \end{split}
\end{equation*}

where for each problem instance
\begin{itemize}
    \item $N^c$ is the total number of parameters to be recovered.
    \item ${\bf \Pi} = (\pi_{i,\alpha} : i\in [n], \alpha \in C_i)$ is the tuple of mixing proportions used for creating the mixture.
    \item ${\bf \hat{\Pi}} = (\hat{\pi}_{i,\alpha} : i\in [n], \alpha \in C_i)$ is the tuple of mixing proportions recovered by our method DIMM.
    \item $Obj()$ is the value of the objective function in $OPT(\epsilon)$ after convergence. 
\end{itemize}

\subsection{ALARM data results}
\label{subsec:ALARM-data-results}
\begin{itemize}
    \item Table \ref{tab:error_metrics} shows a comparison between the true mixture proportions ${\bf \Pi}$ and the proportions ${\bf \hat{\Pi}}$ recovered by our method DIMM, for different problem instances (varying $N_{ivn}$), under the metrics of evaluation defined above.
    
    \begin{table}
\label{}
    \centering
    \resizebox{0.4\textwidth}{!}{\begin{tabular}{ccccc}
    \toprule
    $N_{ivn}$     & \texttt{MSE} & \texttt{MAPE}  & \texttt{MABRE}  & $\Delta(\bm{\Pi},\bm{\hat{\Pi}})$\\
    \midrule 
    0  &  5.32e-7  &  1.02e-4  &  6.92e-3  &  3.70e-4   \\
    5  &  3.65e-5  &  2.32e-3  &  3.09e-2  &  1.09e-2   \\
    9  &  5.26e-5  &  2.99e-3  &  3.40e-2  &  3.71e-2   \\
    17  &  3.78e-5  &  2.61e-3  &  2.58e-2  &  7.63e-3 \\
    34  &  5.19e-5  &  3.29e-3  &  3.46e-2  &  4.33e-2  \\
    68  &  4.22e-5  &  3.48e-3  &  3.46e-2  &  6.37e-2  \\
    \bottomrule
    \end{tabular}}
    \caption{Evaluation of estimated value of mixture proportions $\bm{\hat{\Pi}}$ in terms of various metrics as a function of $N_{ivn}$}
    \label{tab:error_metrics}
\end{table}
    
    \item We also compare our results with that of Expectation-Maximization (EM) based maximum likelihood approach (see Appendix \ref{appendix:EM}). Note that in our problem setup, EM cannot be used as it is since we only get marginals and not the joint distributions. For the sake of this comparison we used samples from the joint, which were used for estimation by the oracle. We compare MAE for both the methods DIMM and EM\footnote{For DIMM, the algorithm is repeated for 60 runs and the best solution in terms of the objective function, is chosen. In case of EM, the solution with the maximum likelihood is chosen out of 50 runs.} in Figure \ref{fig:mae} (in Appendix \ref{appendix:result-plots}) and show that the performance difference between the two methods is negligible when the estimates are good (i.e. large number of samples used for estimates), even though DIMM uses only aggregate information whereas EM requires samples from the joint distribution. Moreover, EM takes over 100 times more time to run as compared to DIMM and is therefore not a scalable solution.
\end{itemize}
\subsection{E-commerce data experiments}
\label{subsec:Ecommerce-data-expts}
We run our DIMM algorithm on the real world e-commerce dataset discussed in section \ref{subsec:e_commerce_data} and provide a qualitative analysis of the obtained parameters. The results are evaluated by drawing a comparison with the solution returned by the EM algorithm (see Appendix \ref{appendix:EM}). We also show convergence of our DIMM method by plotting the objective of DIMM against the number of iterations our solver is run for.
\subsection{E-commerce data results}
\label{subsec:Ecommerce-data-results}
On our real world dataset comprising of total 1122 categories, 54 are identified as the intervened categories with non-zero mixing proportion in the model. This comes after thresholding the estimated parameters at 0.001. The total sum of the mixing proportions of the interventional distributions is obtained as 0.84 implying that the ground network has a contribution of 0.16 in the mixture. The maximum contribution of an intervention is 0.16.

Due to lack of the actual ground truth, we evaluate the above results against the solution returned by the EM algorithm. The mixture proportions uncovered using DIMM are 88\% accurate\footnote{The accuracy is reported after labelling the non-zero mixing proportions as 1 and the others as 0.} with an MSE of 5.4e-5 and MAE of 1.1e-3 with respect to EM.

The curve in Figure \ref{fig:convergence} in Appendix \ref{appendix:result-plots} plots the objective of our optimization problem against the number of iterations it has run for. Our objective converges to a value of 0.097.

\section{Conclusion and future work}
\label{section:conclusions}
We described a problem regarding separating a mixture of distributions resulting from interventions on a causal network. The problem's motivation comes from an application in e-commerce analytics, where an analyst tries to attribute credit/blame to interventions for observed deviations in metrics, with access to aggregate data only. An interesting direction for future work would be to consider more complicated interventions as part of the mixture and prove identifiability. Another more ambitious project would be to improve on the algorithm in \cite{Thiesson1998} after specializing to the case where all DAGs in the mixture correspond to interventions on some unknown causal bayesian network.

%We formally defined the mixture model of interventions and using an example showed that in general identifying mixture proportions from marginals and the ground network was not possible. After making a mild assumption, backed in real world application, we gave an identifiability proof for the mixture proportions. Since in the problem setup, only estimates to marginals of the mixture distribution are known, the identifiability proof cannot be used to compute the mixture proportions and so we design an optimization approach to find the best fit for the mixture proportions. Finally we validated our results using a simulated mixture on the publicly available ALARM network and on data from a large e-commerce business.

%The goal would be to learn mixture proportions and the causal network simultaneously.

% \bibliographystyle{alpha}
\bibliographystyle{aaai}
\bibliography{main}

\clearpage

\appendixpage
\appendix
\section{Proofs from Section \ref{section:problem-statement}}
\label{appendix:proofs}

\begin{proof}[Proof of Lemma \ref{lemma:intersection}]
Proof of the first part is constructive. We can just plug $x_k=0$ in the equations describing line $\mathcal{L}$ for $k\in [K]$. It's easy to see that we get a unique vector $p_k\in \mathbb{R}^K$ on doing so. The second part is almost by definition. We know that $(\pi_{j,\beta_1}, \ldots,\pi_{j,\beta_K})^T$ lies on $\mathcal{L}$. It belongs to one of the $\mathcal{H}_k$'s by Assumption \ref{assumption:identifiability-assumption} and therefore it's inside $\mathcal{P}$.
\end{proof}

\begin{proof}[Proof of Lemma \ref{lemma:uniqueness}]
Assume without loss of generality that $p_1, p_2$ are distinct and both belong to $\mathbb{R}_{\geq 0}^K$. Let $p_i = (p_{i,1}, \ldots, p_{i,K}), i\in [2]$.  Since $p_i \in \mathcal{L}$, we know using equation \ref{equation:solution-line} that for $i\in [2], k\in [K-1]$, 
\[
p_{i,k} = \frac{a_k}{a_K}p_{i,K} + b_k^\prime. 
\]
This can be used to get,
\[
  \begin{bmatrix}
  p_{1,1}-p_{2,1}\\
  .\\
  .\\
  .\\
  p_{1,K} - p_{2,K}
  \end{bmatrix}
   = 
   \lambda \begin{bmatrix}
  \frac{a_1}{a_K}\\
  .\\
  .\\
  \frac{a_{K-1}}{a_K}\\
  1
  \end{bmatrix}
\]
where $\lambda\in \mathbb{R}$. $\lambda$ cannot be $0$, since it'll imply $p_1=p_2$ (which is not the case by assumption). Since $p_i\in \mathcal{H}_i$, we know that $p_{1,1}=0$ and $p_{2,2}=0$. Plugging this in equations above we get,
\[
 -p_{2,1} = \lambda \frac{a_1}{a_K},
\]
\[
 p_{1,2} = \lambda \frac{a_2}{a_K}
\]
Since we have assumed $p_1,p_2\in \mathbb{R}_{\geq 0}^K$, we know that $p_{2,1} \geq 0$ and $p_{1,2} \geq 0$. We also know that $a_1,a_2,a_K$ are all strictly positive. First equation above implies $\lambda <0$ and second implies $\lambda >0$. Hence we arrive at a contradiction. Thus $p_1=p_2$ and there is a unique $k\in [K]$ such that $p_k\in \mathbb{R}_{\geq 0}^K$, completing the proof.
\end{proof}
\section{Preliminaries From Causality}
\label{appendix:preliminaries}
Below we define bayesian networks, interventions, causal bayesian networks and mixture of interventions. We encourage the interested readers to follow up on these definition using the cited literature. 

\begin{definition}[Bayesian Network, Chapter $8$, \cite{Bishop2006}]
Let ${\bf X} = (X_1,\ldots,X_n)$ be a multi-variate random variable. A bayesian network $\mathcal{M} = (G,\mu)$ is:
\begin{itemize}
    \item a directed acyclic graph $G$ on nodes $X_1,\ldots,X_n$ such that the joint probability $P({\bf X})$ factorizes as,
    \begin{equation}
    \label{defn:markov}
        P({\bf X}) = \prod\limits_{i=1}^n P(X_i|Pa(X_i))
    \end{equation}
    where $Pa(X_i)$ are the parents of $X_i$ in $G$, and
    \item parameters $\mu$ which determine the $n$ conditional probabilities $\{P(X_i | Pa(X_i)))\}_{i=1}^n$.
\end{itemize}
\end{definition}
If a joint probability $P$ factors relative to a directed acyclic graph $G$ as in equation \ref{defn:markov}, then we say that $P$ is \emph{markov} with respect to $G$.
\subsubsection{Interventions and do operator}
Let ${\bf X}$ be a multi-variate random variable as in the above definition. Below we define the concept of an intervention distribution. 
\begin{definition}[Intervention]
\label{defn:intervention}
Let $S\subset [n]$ and ${\bf X_S}$ be the tuple of variables in $(X_s)_{s\in S}$. An intervention on ${\bf X_S}$ with value $x_S = (x_s)_{s\in S}$, also denoted as $do({\bf X_S} = x_S)$ sets the value of ${\bf X_S}$ to $x_S$ and leads to the interventional distribution:
\[
P({\bf X} | do({\bf X_S} = x_S)).
\]
It captures the joint distribution when the effect of everything else on ${\bf X_S}$ is nullified and ${\bf X_S}$ is forced to take the value $x_S$. 
\end{definition}
Interventions where $|S|\leq 1$ are called \emph{perfect interventions} \cite{Eaton2007}. Note that in a perfect intervention, either a single variable is intervened and set to some value or none of the variables are intervened.
\section{Optimization Formulation}
\label{appendix:opt-formulation}
We define our optimization problem below. An intuition on how we arrive at this problem follows the definition.
\subsection{Optimization setup}
\label{subsection:optimization-setup}

%is given in Appendix \ref{appendix:optimization-formulation}.

\begin{definition}[$OPT(\epsilon)$]
\label{defn:opt}
Using the ideas and definitions given above, for $\epsilon > 0$, we define our optimization problem $OPT(\epsilon)$.
\[
\min_{{\bf x}} \{\max_{j\in [n]}\|{\bf A_jx_j - \tilde{b}_j}\|_2 + \lambda \|{\bf x}\|_2\}
\]
such that
\[
g_{j, \beta}({\bf x}) = x_{j,\beta} \geq 0 \hspace{2em} \forall j\in [n], \beta\in C_j,
\]
\[
g({\bf x}) = \sum\limits_{j\in [n]}\sum\limits_{\beta\in C_j} x_{j,\beta} \leq 1, \hspace{2em} and
\]
\[
g_j^{\epsilon}({\bf x}) = \epsilon - \min_{k\in[|C_j|]} x_{j,k} \geq 0 \hspace{2em} \forall j\in [n]
\]
\end{definition}
We sketch the idea behind the above optimization problem below. Recall from equation \ref{equation:perfect-intervention-mixture-marginal-notation} that\footnote{Here $p(j,\beta) = P(X_j=\beta), p_{mix}(j,\beta) = P_{mix}(X_j = \beta)$ and $p_{i,\alpha}(j,\beta) = P(X_j = \beta | do(X_i = \alpha))$.}
\begin{equation*}
\begin{split}
  p_{mix}(j, \beta) = \sum\limits_{i\leq j}\sum\limits_{\alpha\in C_i} (p_{i,\alpha}(j, \beta) - p(j, \beta) ) \pi_{i,\alpha} + p(j, \beta)  
\end{split}
\end{equation*}

This can be re-written as:
\[
\sum\limits_{i\leq j}\sum\limits_{\alpha\in C_j}a_{i,\alpha}(j,\beta)\pi_{i,\alpha} = b(j,\beta)
\]
Here $a_{i,\alpha}(j,\beta) = p_{i,\alpha}(j, \beta) - p(j, \beta)$ and $b(j,\beta) = p_{mix}(j,\beta) - p(j,\beta)$. For every $j\in [n]$, by varying $\beta\in C_j$, this gives a system of linear equations satisfied by the ordered tuple $\bm{\pi}_j = (\pi_{i,\alpha} : i\leq j, \alpha\in C_i)$. Let's collect the coefficients of the $\pi_{i,\alpha}$'s in the above equation in a matrix ${\bf A_j}$ and constants in vector ${\bf b_j}$. Therefore for every $j\in [n]$ we have a linear system,
\[
{\bf A_j x_j = b_j}
\]
on variables ${\bf x_j} = (x_{i,\alpha} : i\leq j, \alpha\in C_i)$
such that $\bm{\pi}_j$ is a solution to the system.

Since we only have estimates to the marginals of the mixture distribution, we can only compute estimates ${\bf \tilde{b}_j}$ of ${\bf b_j}$ . This motivates us to try minimizing the objective,
\[
f_j({\bf x_j}) = \|{\bf A_jx_j - \tilde{b}_j}\|_2
\]
where for any vector $v = (v_1,\ldots,v_K)^T \in \mathbb{R}^K$, the norm $\|v\|_2 = \sum\limits_{k=1}^K |v_k|^2$. Since we have $n$ such objectives we take a conservative approach and try minimizing the objective,
\[
\max_{j\in [n]}{f_j({\bf x_j})} = \max_{j\in [n]}\|{\bf A_jx_j - \tilde{b}_j}\|_2
\]

Let ${\bf x}$ denote the ordered tuple of solutions to the full linear system, i.e. ${\bf x} = (x_{j,\beta} : j\in [n], \beta\in C_j)$. Since now we are working with estimates, solutions to the above minimization might not be valid probabilities (which we want them to be). Thus we enforce it using the following linear constraints
\begin{equation*}
    \begin{split}
       & g_{j, \beta}({\bf x}) = x_{j,\beta} \geq 0 \\
       & g({\bf x}) = \sum\limits_{j\in [n]}\sum\limits_{\beta\in C_j} x_{j,\beta} \leq 1
    \end{split}
\end{equation*}

for all $j\in [n]$ and $\beta\in C_j$. We also enforce Assumption \ref{assumption:identifiability-assumption} using the equality constraint:
\[
g_j({\bf x}) = \prod\limits_{\beta\in C_j}x_{j,\beta} = 0.
\]

This constraint will be needed since it guaranteed us identifiability. If we do not impose this constraint, even under perfect computation of marginals, our solution might not be unique.

Constraints $g_j({\bf x}), j\in [n]$ are highly non-linear and therefore hard to manage in iterative solutions. We relax them to $g_j^{\epsilon}({\bf x})$ using a parameter $\epsilon$ such that for $\epsilon=0$ we recover the original constraint. For every $j\in [n]$, define constraint
\[
g_j^{\epsilon}({\bf x}) = \epsilon - \min_{k\in[|C_j|]} x_{j,k} \geq 0.
\]

Due to noise in estimates ${\bf \tilde{b}_j}$ of ${\bf b_j}$, our solutions can over-fit by fitting to this noise. So we add a $L_2$ regularization term to our objective with regularization parameter $\lambda$.

\subsection{Choice of Solver}
\label{subsection:description-slsqp}
In our optimization problem $OPT(\epsilon)$ (see Definition \ref{defn:opt}), objective function is piece-wise quadratic, constraints are either linear or piece-wise linear, so we choose the sequential quadratic programming \cite{Nocedal2006} algorithm to solve the problem iteratively. We use the Sequential Least SQuares Programming (SLSQP) implementation in the scipy library \cite{scipy}. Parameters $\epsilon$ is set to a very small value and $\lambda$ is tuned manually in the range $[0,1]$. Exact values of these parameters are specified while showing the results in section \ref{section:Expts-and-results}. Due to space constraints, we encourage the reader to find more details in the cited literature.
\section{Log-likelihood and EM algorithm}
\label{appendix:EM}

Given independent samples ${\bf D} = \{{\bf x^1, \ldots , x^m}\}$ from the mixture distribution (see section \ref{subsection:notation-and-preliminaries} for definition of the mixture distribution), the log likelihood function can be computed as:
\begin{equation}
    \begin{split}
        \mathcal{L}(\bm{\Pi}) = \log {P[{\bf x^1, \ldots, x^m} | {\mathcal{M}, \bm{\Pi}}]} = \sum\limits_{i=1}^m \log{P[{\bf x^i | {\mathcal{M}, \bm{\Pi}}}]} \\= \sum\limits_{i=1}^m \log{(\sum\limits_{j=1}^n\sum\limits_{\beta\in C_j} \pi_{j,\beta}P_{j,\beta}({\bf x^i})  + \pi_\phi P_\phi({\bf x^i}))}
    \end{split}
\end{equation}
where $\mathcal{M}$ is a causal bayesian network with distribution $P$, $P_{i,\alpha}$ $i\in [n], \alpha\in C_i$ are distributions resulting from interventions on $\mathcal{M}$ (see Definition \ref{defn:intervention}),
$P_\phi$ is the distribution resulting from no intervention and therefore is same $P$, 
${\bm{\Pi}} = \{\pi_{i,\alpha} : i\in \{1,\ldots,n\}, \alpha\in C_i\}$ is the set of all mixing proportions.

Since maximizing the log likelihood does not lead to a closed form analytical solution, we can use an Expectation-Maximization (EM) approach. The parameters of individual networks are already known and the algorithm only needs to compute the optimal w.r.t. parameters in ${\bm{\Pi}}$. A typical EM algorithm for this purpose will involve the following steps (chapter $9$ in \cite{Bishop2006}):
\begin{enumerate}
    \item Choose initial setting $\pi_{i,\alpha}^{old}$ for parameters in ${\bm{\Pi}}$.
    \item \label{item:E-step} {\bf E step: } Evaluate ``responsibilities'' $\gamma_{j, (i,\alpha)}$ denoting the posterior probability that distribution $P_{i,\alpha}$ was responsible for generating data sample ${\bf x^j}$. This can be calculated as:
    \[
    \gamma_{j, (i,\alpha)} = \frac{\pi_{i,\alpha}P_{i,\alpha}({\bf x^j})}{\sum\limits_{i=1}^n\sum\limits_{\alpha\in C_i}\pi_{i,\alpha}P_{i,\alpha}({\bf x^j}) + \pi_\phi P_\phi({\bf x^j})}
    \]
    \item {\bf M step: } Evaluate new mixing proportions
    \[
    \pi_{i,\alpha}^{new} = \frac{\sum_{j=1}^m \gamma_{j, (i,\alpha)}}{m}.
    \]
    \item Evaluate log likelihood using the new parameters. If convergence is not established, return to step \ref{item:E-step}.
\end{enumerate}
\section{Plots from Results in Section \ref{section:Expts-and-results}}
\label{appendix:result-plots}
\begin{figure*}[t]
    \centering
    \begin{subfigure}[b]{0.32\textwidth}
        \centering
        \includegraphics[width =\textwidth]{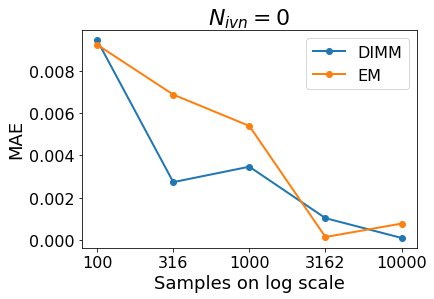}
        % \caption{$N_{ivn} = 0$}
        \label{subfig:mse1}
    \end{subfigure}
        \begin{subfigure}[b]{0.32\textwidth}
        \centering
        \includegraphics[width =\textwidth]{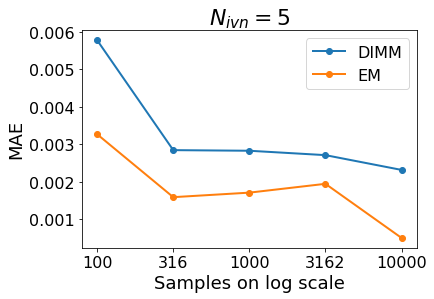}
        % \caption{$N_{ivn} = 5$}
        \label{subfig:mse2}
    \end{subfigure}
        \begin{subfigure}[b]{0.32\textwidth}
        \centering
        \includegraphics[width =\textwidth]{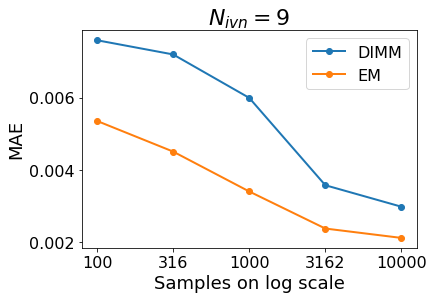}
        % \caption{$N_{ivn} = 9$}
        \label{subfig:mse3}
    \end{subfigure}
        \begin{subfigure}[b]{0.32\textwidth}
        \centering
        \includegraphics[width =\textwidth]{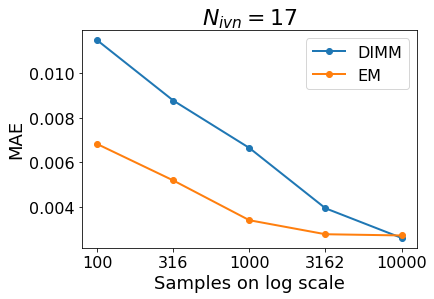}
        % \caption{$N_{ivn} = 17$}
        \label{subfig:mse4}
    \end{subfigure}
        \begin{subfigure}[b]{0.32\textwidth}
        \centering
        \includegraphics[width =\textwidth]{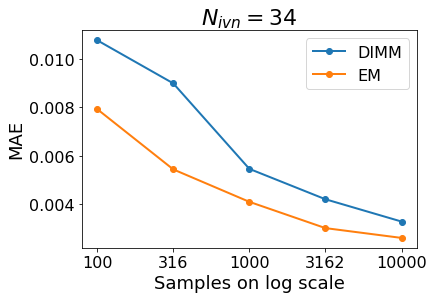}
        % \caption{$N_{ivn} = 34$}
        \label{subfig:mse5}
    \end{subfigure}    
    \begin{subfigure}[b]{0.32\textwidth}
        \centering
        \includegraphics[width =\textwidth]{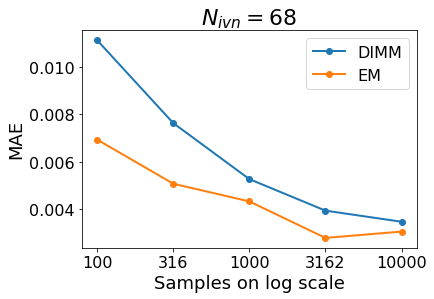}
        % \caption{$N_{ivn} = 68$}
        \label{subfig:mse6}
    \end{subfigure}
    
    \caption{Comparison of MAE in estimation of $\bm{\Pi}$ by DIMM and EM as a function of number of samples $N_s$ at different values of $N_{ivn}$}
    \label{fig:mae}
\end{figure*}

\begin{figure}[H]
\centering
\includegraphics[width=0.4\textwidth]{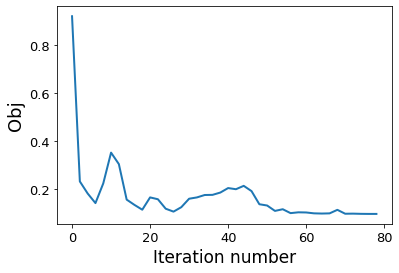}
\caption{Convergence of the objective function using DIMM on e-commerce data}
\label{fig:convergence}
\end{figure}

\end{document}